\documentclass[11pt]{article}
\usepackage[margin=1in]{geometry}
\usepackage{cite}
\usepackage{graphicx,xcolor}
\usepackage{amsmath,amssymb}
\usepackage{booktabs}
\usepackage{siunitx}
\DeclareSIUnit{\px}{px}
\usepackage{url}
\usepackage{subcaption}
\usepackage{enumitem}
\usepackage{multirow}
\usepackage{numprint}
\usepackage{threeparttable}
\usepackage{float}
\usepackage[hidelinks]{hyperref}
\usepackage{authblk}

\def\BibTeX{{\rm B\kern-.05em{\sc i\kern-.025em b}\kern-.08em
    T\kern-.1667em\lower.7ex\hbox{E}\kern-.125emX}}

\title{Lightweight Machine Learning-Driven Monocular Sidewalk Path Extraction for Embedded Micromobility Navigation}

\author[1]{Lkhanaajav Mijiddorj}
\author[1]{Yang Yan}
\author[1]{Tyler Beringer}
\author[1]{Bilguunzaya Mijiddorj}
\author[1]{Alex N. Ho}
\author[2]{Bin Xu}
\author[1]{Binbin Weng\thanks{Corresponding author: Binbin Weng (binbinweng@ou.edu).}}
\affil[1]{School of Electrical and Computer Engineering, University of Oklahoma, Norman, OK 73019, USA}
\affil[2]{School of Aerospace and Mechanical Engineering, University of Oklahoma, Norman, OK 73019, USA}
\date{}

\begin{document}

\maketitle

\begin{abstract}
Sidewalk-scale path extraction demands perception and planning that run reliably on compact, low-power hardware in cluttered, map-sparse environments. We present a monocular vision pipeline for sidewalk path extraction in micromobility systems that progresses through three design iterations---from a skeleton-graph baseline through distance-transform corridor planning to a lightweight image-space architecture---and provides a systematic comparison of five path-planning methods across both bird's-eye-view (BEV) and image-space domains. A compact SegFormer-B0 student model, trained with a semi-supervised teacher-student framework using OneFormer Swin-L pseudo-labels, achieves a hand-annotated IoU of 0.946 at 11.7~ms per frame, improving over the baseline checkpoint (IoU 0.758, 18.9~ms). In a controlled planner comparison on 32 hand-labeled frames, image-space midpoint planning achieves the lowest lateral center error (14.3~px) at 2.2~ms---a 421$\times$ speedup over BEV distance-transform planning (926.8~ms, 65.0~px center error)---while maintaining comparable mask-path alignment (98.5\% versus 98.6\%). A full-video replay across six campus sequences (22,679 frames) confirms that the improved segmentation reduces temporal instability from 1.46\% to 0.33\% and increases template-path availability from 73.7\% to 79.3\%. We further show that BEV-only path extraction is fragile in monocular settings: in one profiled run, 99.3\% of frames produced no valid BEV path. The final recommended architecture---image-space midpoint primary, image-space distance-transform fallback, and BEV reserved for visualization---runs the full perception-to-path stack in under 50~ms per frame on CPU, making it suitable for embedded pedestrian-speed micromobility systems.
\end{abstract}


\section{Introduction}
Autonomous micro-mobility is beginning to move from controlled demonstrations to everyday sidewalks. Delivery carts, assistive scooters, and other pedestrian-speed robots must operate among pedestrians, signs, bicycles, vegetation, and irregular curb geometry, all under tight payload, power, and cost constraints. Sidewalks further complicate perception: widths vary, surfaces mix bricks and concrete, markings are inconsistent, and lighting changes rapidly under trees and buildings. High-definition pedestrian maps are scarce and GNSS is unreliable near canopies and urban canyons. In this setting, a vision-first navigation pipeline that is robust, interpretable, and efficient enough for single-board computers is essential.

We adopt a deliberately simple, geometry-aware approach for local sidewalk path extraction. A lightweight monocular segmentation model extracts the traversable sidewalk region. From this mask, we extract navigation paths using one of two geometric domains: a bird's-eye view (BEV) obtained via planar homography, or direct image-space boundary analysis. We systematically compare five planning methods across both domains and find---perhaps surprisingly---that simple image-space geometry consistently outperforms the more complex BEV pipeline in both accuracy and latency for this monocular sidewalk setting.

This direction complements several active research threads. End-to-end steering policies trained on rich sensors can perform well, but they are difficult to interpret and often assume GPU-class hardware~\cite{e2e_nav_policies, viteri2024}. Dense monocular BEV reconstruction and diffusion-based free-space methods push accuracy but remain challenging to deploy on low-power platforms~\cite{zhao2024bev, gupta2025diffusion}. Classic semantic-mask corridor following is efficient in structured environments but does not explicitly expose intersection topology~\cite{orchard_seg_nav, row_nav_survey}. In contrast, our approach is monocular, transparent, and light enough for CPU-only inference, while providing an empirical comparison of planning-domain trade-offs that is absent from prior sidewalk navigation work.

\begin{figure}[t]
  \centering
  \begin{subfigure}{0.49\linewidth}
    \centering
    \includegraphics[width=\linewidth]{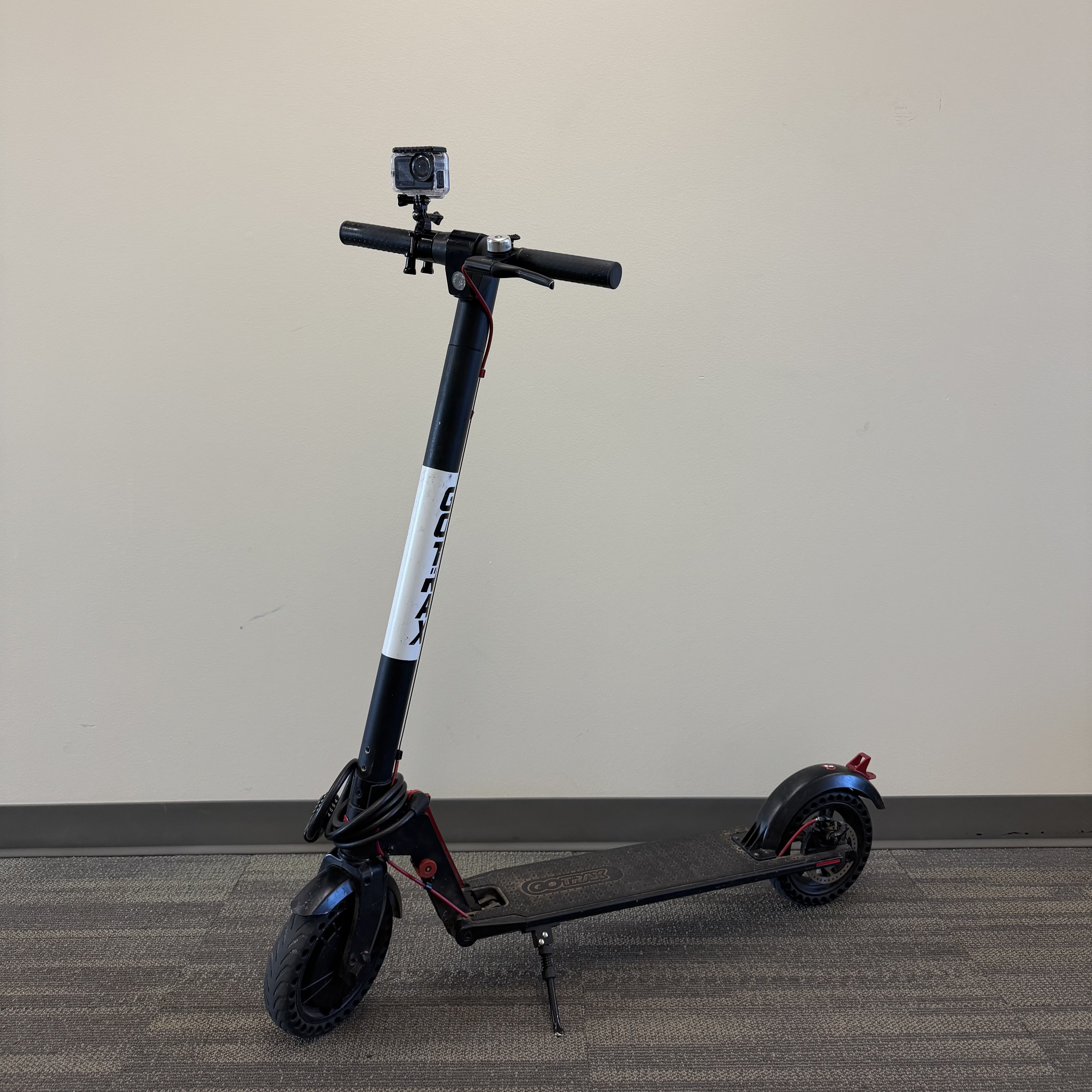}
    \caption{Scooter platform with mounted forward-facing camera.}
    \label{fig:scooter_hw1}
  \end{subfigure}
  \hfill
  \begin{subfigure}{0.49\linewidth}
    \centering
    \includegraphics[width=\linewidth]{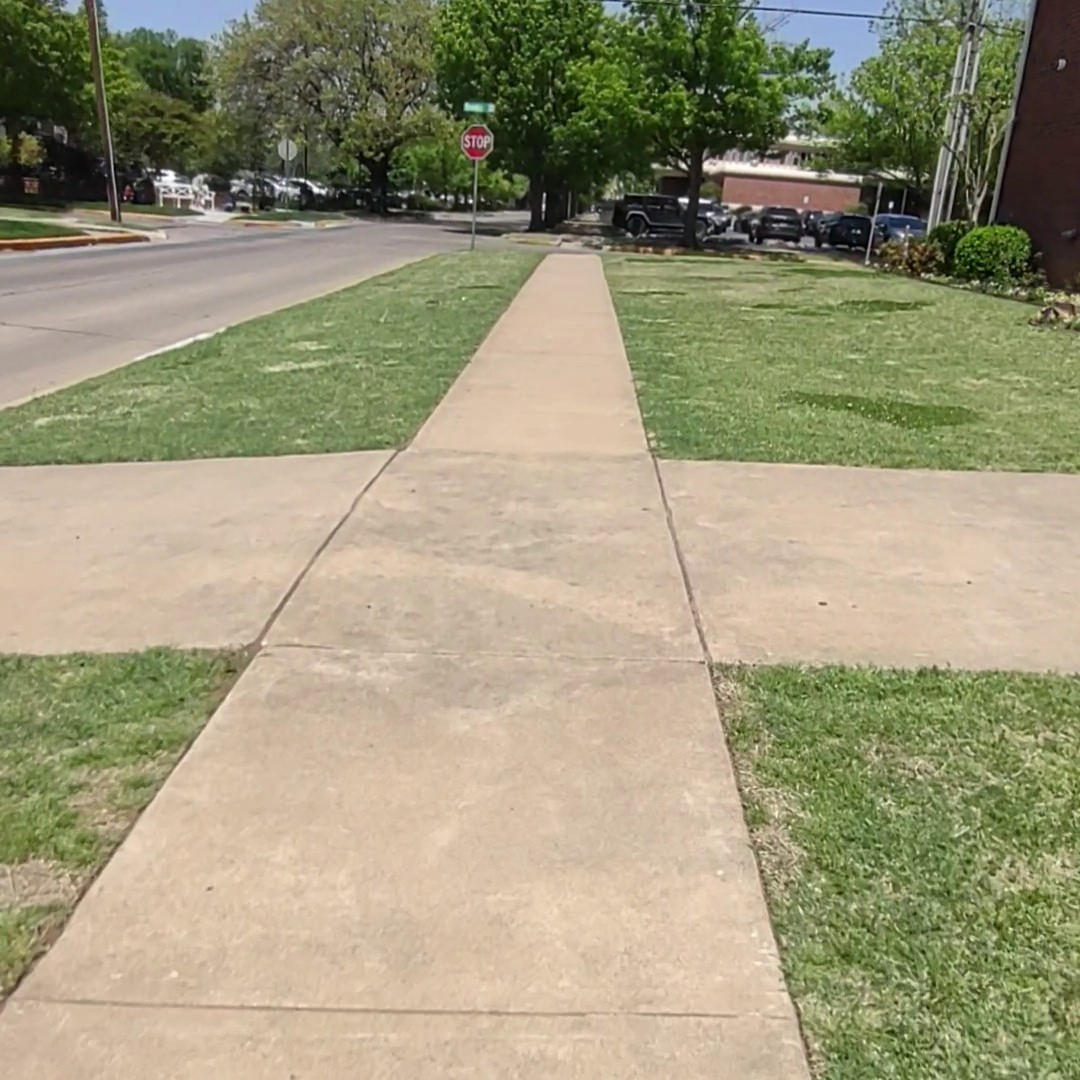}
    \caption{Example first-person view from the scooter-mounted camera.}
    \label{fig:scooter_hw2}
  \end{subfigure}
  \caption{Hardware context for this study. (a) Electric scooter platform used for data collection, with a forward-facing monocular camera rigidly mounted to the handlebar stem. (b) Sample sidewalk scene as seen by the camera; all experiments are conducted offline on video recorded from this rig.}
  \label{fig:scooter_hw}
\end{figure}

In this work, we focus on validating the perception and path-generation pipeline offline. Using video recorded from a forward-facing camera mounted on an electric scooter (Fig.~\ref{fig:scooter_hw}), we apply the full pipeline frame by frame and evaluate the resulting paths without closing the loop on the vehicle. This isolates perception and planning, enabling detailed analysis of segmentation quality, planner accuracy, runtime efficiency, and the BEV-versus-image-space trade-off.

\textbf{Contributions.}
\begin{itemize}[leftmargin=*,nosep]
  \item A modular monocular pipeline for sidewalk path extraction
that is transparent, lightweight, and designed for embedded
hardware, progressing through three design iterations with
backward-compatible improvements.
  \item A semi-supervised segmentation training recipe using a high-capacity OneFormer Swin-L teacher to generate pseudo-labels for a compact SegFormer-B0 student, achieving hand-annotated IoU of 0.946 at \SI{11.7}{ms} on CPU.
  \item A systematic comparison of five path-planning methods across BEV and image-space domains, demonstrating that image-space midpoint planning achieves a 421$\times$ speedup over BEV distance-transform planning with lower lateral error.
  \item Evidence that BEV-only planning is fragile in monocular sidewalk settings, with 99.3\% of frames failing to produce a valid BEV path in one profiled sequence, motivating image-space planning as the primary domain.
  \item A comprehensive offline evaluation protocol covering segmentation quality, planner accuracy, temporal stability, and runtime analysis across six campus video sequences totaling over \numprint{22000} frames.
\end{itemize}

\noindent\textbf{Organization.} We review related work, describe the pipeline architecture (segmentation, BEV projection, and five planning methods), define evaluation metrics, present results including the planner comparison study, and conclude with limitations and directions for closed-loop deployment.

\section{Related Work}
We organize prior work into seven threads and emphasize what each direction offers in practice---what runs fast, what transfers, and what stays explainable.

\paragraph{Target-driven and end-to-end navigation.}
Target-driven systems in structured indoor settings combine monocular depth and segmentation to follow objects without explicit mapping~\cite{zhu2017target,machkour2023}. For sidewalks, end-to-end policies trained on RGB-D can steer directly from raw streams~\cite{viteri2024, e2e_nav_policies}. The upside is simplicity at runtime: one model, one policy. The downside is limited interpretability when conditions shift from training, and many solutions assume GPU-class compute. Our pipeline trades some expressivity for a modular stack that can be inspected stage by stage.

\paragraph{Dense monocular bird's-eye reconstruction.}
Monocular BEV models that lift features or regress dense warps produce striking top-down reconstructions~\cite{zhao2024bev}. On modern GPUs, they shine. On single-board computers, memory and latency are a stretch. A calibrated homography is less glamorous but predictable; for near-planar sidewalk patches and short look-ahead, it delivers a stable metric frame without the overhead of full 3D. However, as we show in Section~\ref{sec:bev_fragility}, even a carefully calibrated homography can be fragile when the monocular viewpoint provides insufficient mask coverage for the BEV domain.

\paragraph{Segmentation for drivable/free space.}
Two dominant pushes drive perception: accuracy via ensembles and efficiency via compact backbones. Ensemble approaches can drive mIoU higher~\cite{shihab2024}. Lightweight networks (SegFormer-B0, MobileNet variants) bring latency and memory to embedded levels~\cite{segformer, edge_optimized_seg, twinlitenet2024}. Our approach carries the segmentation mask into geometry and planning, with the goal of a complete path from pixels to motion.

\paragraph{Generative free-space priors and diffusion.}
Diffusion-based free-space predictors can hallucinate plausible corridors when the signal is weak~\cite{gupta2025diffusion}. For an embedded scooter, the footprint and tuning complexity are still a mismatch. We prefer deterministic geometry that behaves consistently.

\paragraph{Corridor following and agricultural rows.}
The agricultural community has a long history of segmentation-driven row navigation~\cite{orchard_seg_nav, row_nav_survey}. Histogram-of-columns minima are fast and reliable when the world is a corridor. Sidewalks inherit the corridor idea but add junctions, driveways, and uneven edges.

\paragraph{Skeletons and mid-level geometry.}
Learned skeletonization can reduce planning cost~\cite{flores2025skeleton}. Classic thinning (Zhang--Suen~\cite{zhang_suen}, Guo--Hall~\cite{guo_hall}) remains attractive in embedded contexts. We adopt classical thinning as one of our planning baselines and compare it against distance-transform and image-space alternatives.

\paragraph{Image-space planning and distance transforms.}
An alternative to BEV-based planning is to operate directly in image coordinates. Per-row boundary midpoint extraction and distance-transform maxima can produce smooth centerlines without the distortion and coverage issues inherent in homography-based BEV projection. While image-space methods sacrifice metric ground-plane reasoning, they avoid the failure modes we document in BEV planning and run at substantially lower latency. This trade-off has received limited attention in sidewalk navigation literature, motivating our systematic comparison.

\section{Methodology}
\label{sec:method}

\begin{figure}[t]
  \centering
  \includegraphics[width=\textwidth]{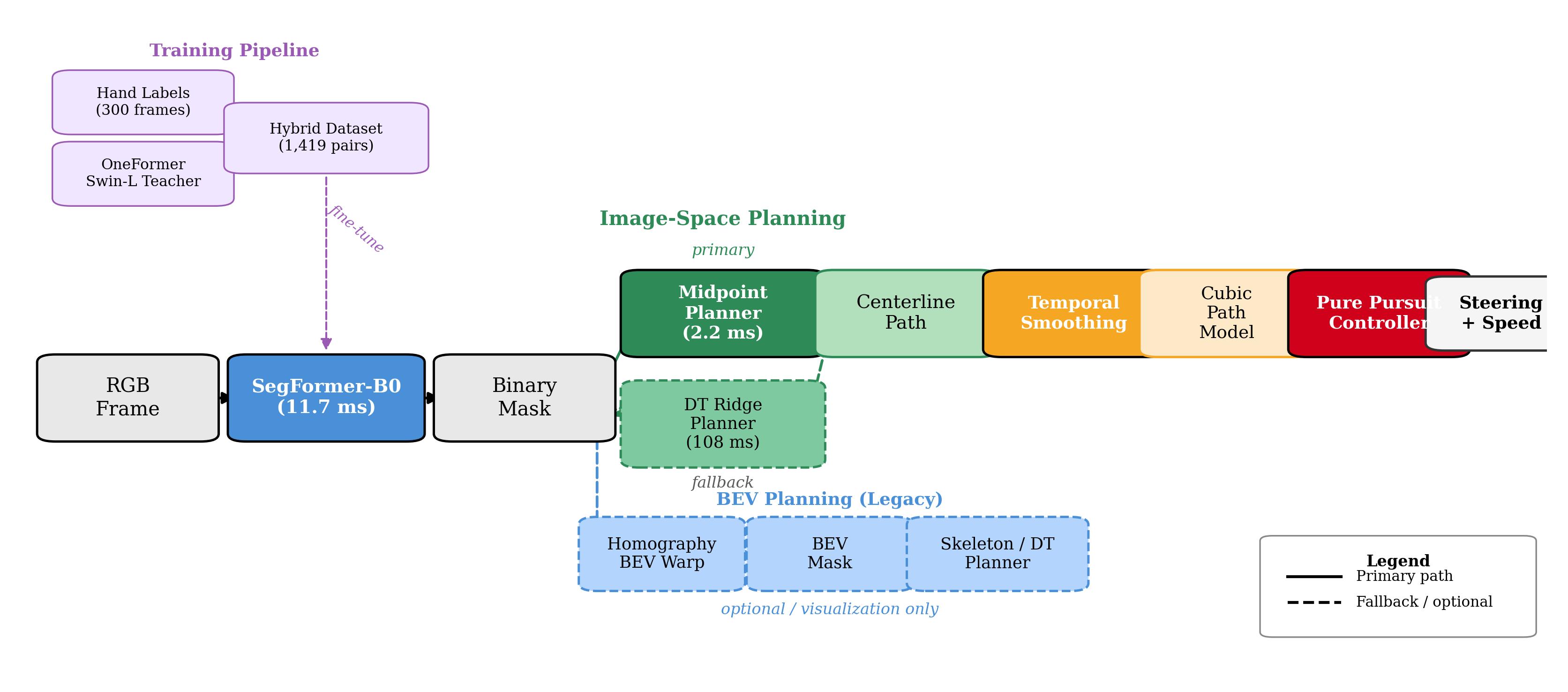}
  \caption{Overview of the sidewalk navigation pipeline. Hand-annotated and pseudo-labeled data are merged to train a student segmentation model, followed by geometric processing to extract traversable paths. The system supports both BEV-domain and image-space planning; our experiments compare five methods across both domains.}
  \label{fig:pipeline_diagram}
\end{figure}

This work proposes a vision-based pipeline for sidewalk perception and local path extraction on sidewalk-following micromobility platforms using only monocular RGB input. As illustrated in Fig.~\ref{fig:pipeline_diagram}, the architecture maps camera frames to controller-ready waypoints through modular stages: semantic segmentation, optional BEV projection, path planning (five methods compared), and waypoint smoothing. The system evolved through three design iterations, each backward-compatible with the previous, enabling controlled comparison.

\subsection{Segmentation Module and Supervision Strategy}
\label{sec:segmentation}

The first stage performs pixel-wise segmentation of sidewalk regions from monocular RGB input. This module is critical for all downstream steps and must balance segmentation quality with real-time efficiency on embedded systems.

We adopt \textbf{SegFormer}~\cite{segformer} as our backbone architecture due to its strong performance on urban benchmarks and adaptability across model scales. For deployment, we use \textbf{SegFormer-B0} (3.7M parameters) for its low latency and memory footprint.

\paragraph{Teacher--Student Supervision.}
Small models trained on limited data often produce noisy masks. To mitigate this, we employ a semi-supervised teacher--student framework. We explored two teacher architectures across our design iterations:

\begin{itemize}[nosep]
    \item \textbf{Iteration~1:} A SegFormer-B2 teacher (24M parameters) fine-tuned on 300 hand-labeled frames, generating pseudo-labels on 2,300 unlabeled frames.
    \item \textbf{Iteration~2:} A high-capacity \textbf{OneFormer Swin-L} teacher~\cite{oneformer} pre-trained on ADE20K, generating dense binary pseudo-labels from up to 1,419 unlabeled frames across 10 campus video sequences.
\end{itemize}

For each unlabeled image $x \in D_U$, the teacher predicts pixel-wise logits $z_T = f_T(x)$, converted to probabilities via softmax: $p_T = \sigma(z_T)$. A confidence threshold $\tau = 0.60$ produces binary masks:
\[
\tilde{y}(u) =
\begin{cases}
1,& \text{if } p_T(u) \ge \tau \\
0,& \text{otherwise}
\end{cases}
\quad \text{with } u \in \Omega
\]
Pseudo-labels are further cleaned with connected-component filtering and morphological smoothing.

\paragraph{Hybrid Training Dataset.}
The training set $\mathcal{D} = D_L \cup D_P$ merges pseudo-labeled frames with hand-labeled frames. In frames where both are available, hand labels take precedence. Our best-performing student was trained on 1,419 image--mask pairs (699 from earlier sequences plus 720 newly extracted frames from four additional campus videos).

\noindent\textbf{Loss Function.}
The student model $f_S$ is optimized using a weighted cross-entropy loss and a Dice loss:
\begin{equation}
\mathcal{L} =
\mathcal{L}_{\mathrm{WCE}} +
\mathcal{L}_{\mathrm{Dice}},
\end{equation}
where the class weights in $\mathcal{L}_{\mathrm{WCE}}$ are computed from the training split.

\begin{table}[H]
\centering
\caption{Known training settings of the SegFormer-B0 student model.}
\label{tab:training_config}
\scriptsize
\renewcommand{\arraystretch}{1.05}
\begin{tabular}{@{}p{0.40\linewidth}p{0.54\linewidth}@{}}
\hline
\textbf{Setting} & \textbf{Value} \\
\hline
Student model & SegFormer-B0 \\
Teacher model & OneFormer Swin-L pretrained on ADE20K \\
Input resolution & $640 \times 360$ \\
Pseudo-label confidence threshold & 0.60 \\
Optimizer & AdamW \\
Initial learning rate & $5 \times 10^{-5}$ \\
Weight decay & $1 \times 10^{-4}$ \\
Batch size & 4 \\
Training epochs & 10 \\
Random seed & 1337 \\
Loss function & Weighted cross-entropy + Dice loss \\
Loss weights & $\lambda_{\mathrm{WCE}}=1$, $\lambda_{\mathrm{Dice}}=1$ \\
Learning-rate schedule & 10\% linear warm-up followed by cosine decay \\
Data augmentation & Horizontal flip, color jitter, and Gaussian blur \\
Data-loader workers & 2 \\
Training framework & Python 3.11.9; PyTorch 2.10.0+cu128 \\
Training hardware & NVIDIA GeForce RTX 5070 \\
\hline
\end{tabular}
\end{table}

Additional dataset-specific settings, including the number of training pairs and class weights, are reported with the corresponding training split.

\paragraph{Training Progression.}
Table~\ref{tab:training_progression} summarizes the segmentation model iterations. The OneFormer teacher with expanded data yields the highest validation IoU (0.960), while the hand-annotated evaluation (Section~\ref{sec:res_seg}) confirms a large improvement over the baseline (IoU 0.758~$\to$~0.946).

\begin{table}[h]
\centering
\caption{Segmentation model training progression. All students are SegFormer-B0; validation IoU is on held-out splits.}
\label{tab:training_progression}
\begin{threeparttable}
\begin{tabular}{lccc}
\toprule
Iteration & Teacher & Train Pairs & Val IoU \\
\midrule
Baseline & SegFormer-B2 & 300 + 2,300 & 0.758\tnote{$\dagger$} \\
OneFormer (4 vid) & OneFormer Swin-L & 400 & 0.944 \\
OneFormer (mixed) & OneFormer Swin-L & 1,419 & \textbf{0.960} \\
\bottomrule
\end{tabular}
\begin{tablenotes}
\small
\item[$\dagger$] Hand-annotated IoU (external evaluation, not internal validation split).
\end{tablenotes}
\end{threeparttable}
\end{table}

\subsection{Resolution Trade-Off Analysis}
\label{sec:resolution}

\begin{figure}[t]
  \centering
  \includegraphics[width=\textwidth]{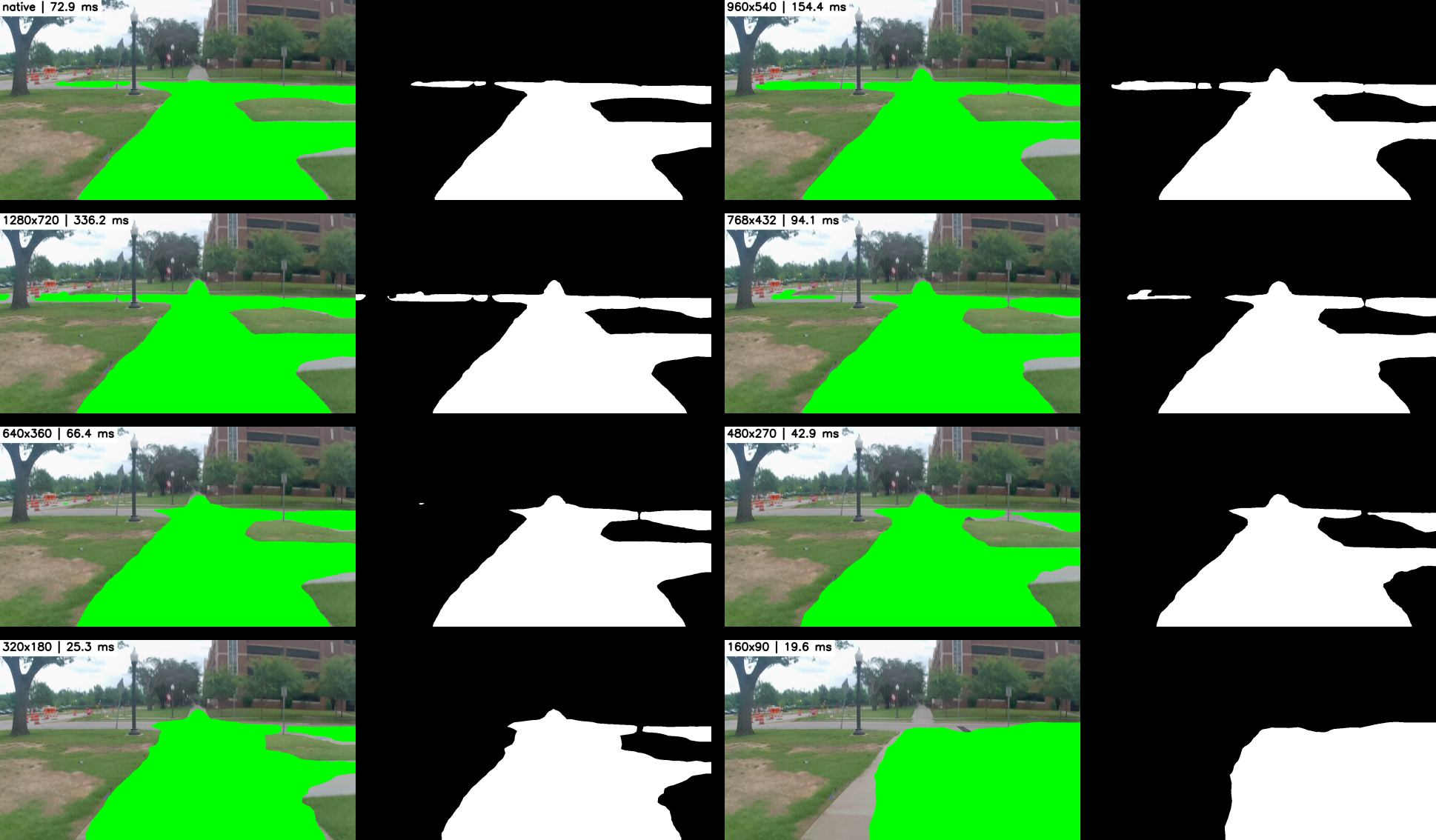}
  \caption{Resolution sweep for SegFormer-B0. Each column shows input overlay (left) and resulting segmentation mask (right). Higher resolutions improve edge detail but increase latency.}
  \label{fig:segformer_compare}
\end{figure}

To balance accuracy and real-time performance, we conducted a resolution sweep using SegFormer-B0 on CPU. Table~\ref{tab:segformer_fps} reports latency and throughput, and Fig.~\ref{fig:segformer_compare} shows the qualitative effect on mask quality. Lower resolutions ($320{\times}180$ and below) produced masks that were overly coarse, fragmenting continuous sidewalk segments. Higher resolutions ($960{\times}540$+) incurred prohibitive latency. We selected $640{\times}360$ as a practical operating point: accurate and consistent segmentation at sub-\SI{50}{ms} inference.

\begin{table}[h]
\centering
\caption{SegFormer-B0 inference latency and throughput at various input resolutions (CPU-only).}
\label{tab:segformer_fps}
\begin{tabular}{lrr}
\toprule
Resolution & Latency (ms) & FPS \\
\midrule
$160{\times}90$         & 19.84 & 50.40 \\
$320{\times}180$        & 25.35 & 39.45 \\
$480{\times}270$        & 42.05 & 23.78 \\
$640{\times}360$        & 46.00 & 21.74 \\
$768{\times}432$        & 93.55 & 10.69 \\
native $540{\times}540$ & 72.80 & 13.74 \\
$960{\times}540$        & 152.45 & 6.56 \\
$1280{\times}720$       & 352.40 & 2.84 \\
\bottomrule
\end{tabular}
\end{table}

\subsection{Bird's-Eye View Projection}
\label{sec:bev}

To transform the segmented mask into a spatially consistent top-down representation, we apply a planar homography. This produces a BEV binary mask in which pixel distances approximate ground distances.

Let $p = [u, v, 1]^\top$ be a homogeneous image pixel and $p' = [x, y, 1]^\top$ its BEV coordinate. The mapping is:
\[
p' \sim H p
\]
where $H$ is estimated via four point correspondences between the camera image and a flat ground reference. The input mask $\widehat{M}_\text{img} \in \{0, 1\}^{h \times w}$ is warped to BEV:
\[
\widehat{M}_\text{bev}(x, y) = \widehat{M}_\text{img}\left( \pi(H^{-1} [x, y, 1]^\top) \right)
\]
where $\pi$ denotes dehomogenization.

The resulting $\widehat{M}_\text{bev}$ is aligned with the robot's reference frame, with the ego position at the bottom-center and forward motion along the vertical axis. This transformation provides metric-scale reasoning but assumes a flat plane and fixed camera pose---an assumption we test empirically in Section~\ref{sec:bev_fragility}.

\subsection{BEV Mask Refinement}
\label{sec:mask_refinement}

The raw BEV mask often contains noise: thin spurs, disconnected blobs, and boundary artifacts from segmentation errors amplified by the perspective warp. We apply a multi-stage cleanup:

\begin{enumerate}[nosep]
    \item \textbf{Morphological closing and opening} with configurable kernel sizes suppress small gaps and spurs.
    \item \textbf{Connected-component filtering} retains only the largest connected region touching the ego anchor row, ensuring the mask represents the sidewalk ahead of the scooter rather than disconnected patches.
    \item \textbf{Enhanced refinement} (optional): flood-fill hole filling, Gaussian boundary smoothing, and distance-transform-based component selection that preferentially retains the component with the highest ego clearance.
\end{enumerate}

These stages are individually toggleable via configuration flags, allowing controlled ablation.

\subsection{Path Planning Methods}
\label{sec:planners}

We compare five path planning methods spanning two geometric domains: three operating in BEV space and two in image space. All methods receive a binary sidewalk mask and produce a centerline trajectory.

\subsubsection{BEV Skeleton-Graph Planner}
\label{sec:skeleton_planner}

The Guo--Hall parallel thinning algorithm~\cite{guo_hall} reduces the BEV mask to a 1-pixel-wide skeleton preserving topological structure. From this skeleton, we construct an undirected graph $\mathcal{G} = (\mathcal{V}, \mathcal{E})$ where each skeleton pixel is a node and 8-neighbor connections form edges. Two-stage pruning (length-based and component filtering) suppresses spurious branches. Dijkstra's algorithm enumerates candidate paths from the ego node $v_0$, each scored by:
\[
C(p_j) = \alpha \cdot \text{Curvature}(p_j) + \beta \cdot \text{LateralShift}(p_j)
\]
This method explicitly exposes junction topology but is computationally expensive due to per-pixel graph construction.

\subsubsection{BEV Distance-Transform Ridge Planner}
\label{sec:dt_planner}

Rather than skeletonize, this method computes the Euclidean distance transform (EDT) of the BEV mask and traces the ridge of maximum clearance. The cost field $c(x,y) = 1/(d(x,y) + \varepsilon)^\alpha$ (where $d$ is the EDT value, $\varepsilon = 0.5$, $\alpha = 1.5$) is minimized via Dijkstra's algorithm to find the globally maximum-clearance path. The resulting centerline is smoothed with a Savitzky--Golay filter. This approach avoids the noise sensitivity of skeletonization but inherits the full cost of BEV computation plus the EDT and graph search.

\subsubsection{BEV Template Arc Planner}
\label{sec:template_planner}

A bank of predefined arc templates (straight, gentle-left, gentle-right, sharp-left, sharp-right) is scored against a corridor extracted from the DT ridge. Each template is evaluated by its containment within the segmented corridor and its smoothness. The highest-scoring template that passes a confidence gate becomes the planned path. This method trades generality for speed and interpretability: the arc bank is fixed and small, making runtime predictable. When no template passes the confidence gate, the system falls back to the DT ridge planner.

\subsubsection{Image-Space Midpoint Planner}
\label{sec:img_midpoint}

This method operates directly on the image-plane segmentation mask $\widehat{M}_\text{img}$, bypassing BEV entirely. For each row $y$ of the mask, the left and right sidewalk boundaries are identified, and the midpoint $x_\text{mid}(y) = (x_L(y) + x_R(y))/2$ is computed. Rows with fewer than a minimum number of road pixels are skipped. The resulting per-row midpoints are smoothed with a Savitzky--Golay filter to produce a continuous centerline. This method is the fastest (\SI{2.2}{ms}) and produces the lowest lateral center error in our experiments.

\subsubsection{Image-Space Distance-Transform Planner}
\label{sec:img_dt}

The EDT is computed on the image-plane mask, and the maximum-distance ridge is traced using dynamic programming with a lateral drift constraint. A per-row forward pass selects the column with minimum cost (inverse EDT) subject to a maximum lateral step per row, producing a path that tracks the widest part of the sidewalk corridor. This method is more robust than midpoint extraction on masks with irregular boundaries or partial gaps but slower (\SI{108.1}{ms}).


\subsection{Temporal Smoothing}
\label{sec:temporal_smoothing}

Frame-by-frame path extraction can produce jittery trajectories due to segmentation noise. We apply two optional temporal filters:

\paragraph{Path temporal smoothing.}
An exponential moving average (EMA) is applied to the cubic polynomial coefficients of the fitted path. The smoothing factor $\alpha$ is adaptive: higher confidence paths receive less smoothing ($\alpha \to 0.85$), while low-confidence paths receive more ($\alpha \to 0.35$). A topology-change detector resets the filter when the path shape changes abruptly (coefficient jump exceeds a threshold), preventing the filter from averaging across distinct path segments.

\paragraph{Heading temporal smoothing.}
A circular EMA filter on the heading angle handles the $\pm 180^\circ$ wraparound discontinuity. The filter resets when the heading delta exceeds $45^\circ$, allowing rapid response to sharp turns while suppressing noise on straight segments.

\subsection{Evaluation Metrics}
\label{sec:metrics}

We evaluate the pipeline offline on held-out campus video and define four quantitative metrics.

\paragraph{Segmentation IoU.}
Intersection-over-union between the predicted binary mask and hand-annotated ground truth, computed per-frame and averaged.

\paragraph{Lateral path-centering error.}
For each sample point along the selected path, we measure the lateral deviation from the geometric centerline of the sidewalk mask. We report mean and standard deviation in pixels (convertible to meters via the known BEV scale).

\paragraph{Mask--path alignment (inside-GT ratio).}
The fraction of path pixels that fall within the ground-truth sidewalk mask:
\[
\text{Alignment} = \frac{L_{\text{in}}}{L_{\text{total}}} \times 100\%
\]

\paragraph{Temporal stability.}
Frame-to-frame IoU between consecutive cleaned masks, with frames below a threshold flagged as ``unstable.'' The unstable rate captures temporal flicker caused by segmentation noise.

\section{Results}
\label{sec:results}

We present results organized by pipeline stage, followed by the planner comparison study and runtime analysis. All experiments use campus video collected at the University of Oklahoma, featuring curved sidewalks, T-junctions, shadows, surface changes, and pedestrian traffic. Six video sequences totaling \numprint{22679} frames are used for full-video evaluation; 32 hand-annotated frames with ground-truth masks serve as the planner comparison benchmark.

\begin{figure}[t]
\centering
\begin{subfigure}[b]{0.32\linewidth}
\includegraphics[width=\linewidth]{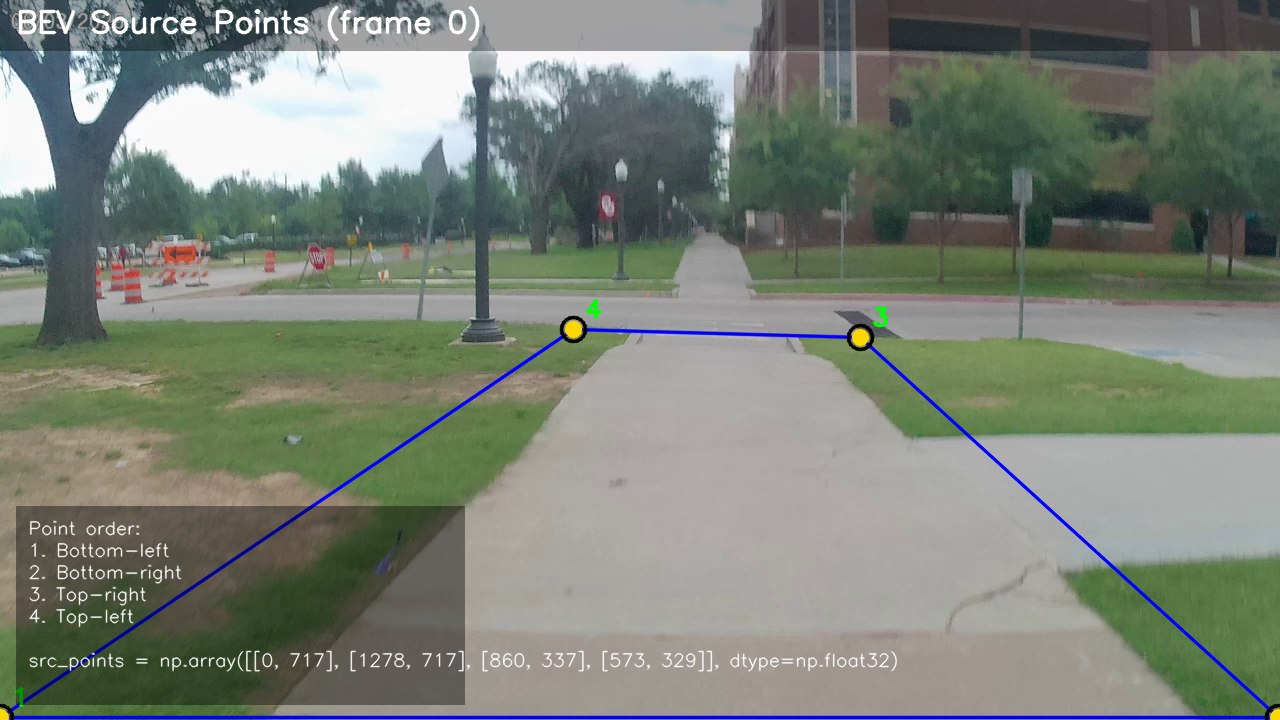}
\caption{RGB input}
\end{subfigure}\hfill
\begin{subfigure}[b]{0.32\linewidth}
\includegraphics[width=\linewidth]{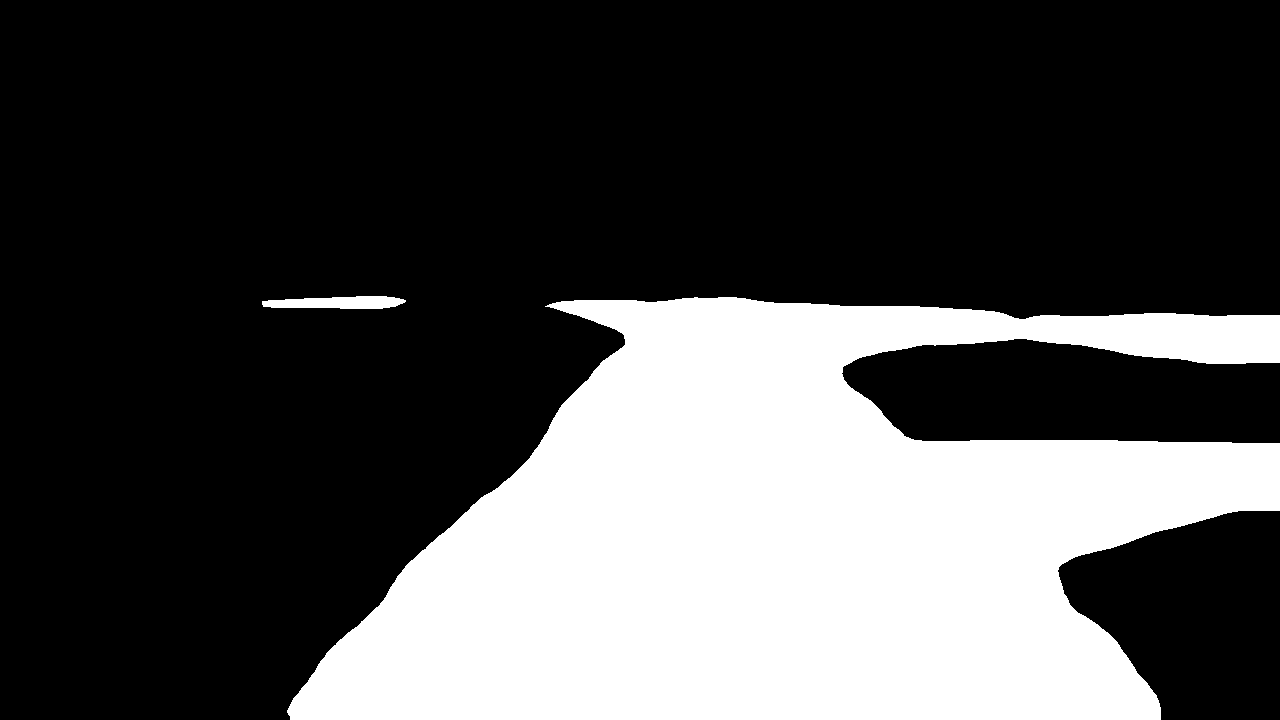}
\caption{Raw segmentation}
\end{subfigure}\hfill
\begin{subfigure}[b]{0.32\linewidth}
\includegraphics[width=\linewidth]{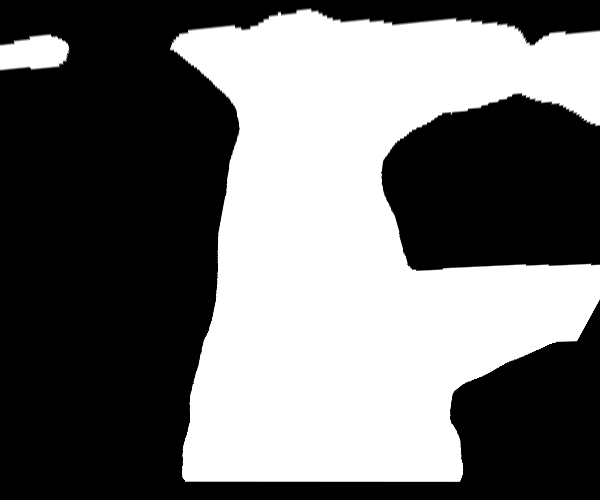}
\caption{BEV projection}
\end{subfigure}
\caption{Segmentation pipeline. (a)~RGB input, (b)~raw sidewalk mask in image plane, (c)~BEV projection of cleaned mask. Morphological filtering and connected-component analysis suppress spurs and disconnected blobs.}
\label{fig:segmentation_stage}
\end{figure}

\subsection{Segmentation Improvement}
\label{sec:res_seg}

Fig.~\ref{fig:segmentation_stage} illustrates the segmentation pipeline stages, and Table~\ref{tab:seg_comparison} compares the baseline and improved segmentation models on 32 hand-annotated frames sampled across multiple campus videos. The improved model (OneFormer Swin-L teacher, threshold 0.60) raises IoU from 0.758 to 0.946 while simultaneously reducing inference time from \SI{18.9}{ms} to \SI{11.7}{ms}---a 38\% speedup. Precision and recall both improve substantially, indicating better mask completeness and fewer false positives.

\begin{table}[h]
\centering
\caption{Segmentation quality on 32 hand-annotated frames. ``Baseline'' uses the original SegFormer-B2 teacher; ``Candidate'' uses the OneFormer Swin-L teacher with optimized threshold.}
\label{tab:seg_comparison}
\begin{tabular}{lccccr}
\toprule
Model & IoU & Prec. & Recall & F1 & ms \\
\midrule
Baseline      & 0.758 & 0.910 & 0.835 & 0.851 & 18.9 \\
Candidate     & \textbf{0.946} & \textbf{0.983} & \textbf{0.962} & \textbf{0.972} & \textbf{11.7} \\
Cand.+confhold & 0.903 & 0.987 & 0.914 & 0.947 & 13.1 \\
\bottomrule
\end{tabular}
\end{table}

\begin{figure}[!t]
  \centering
  \includegraphics[width=0.96\textwidth]{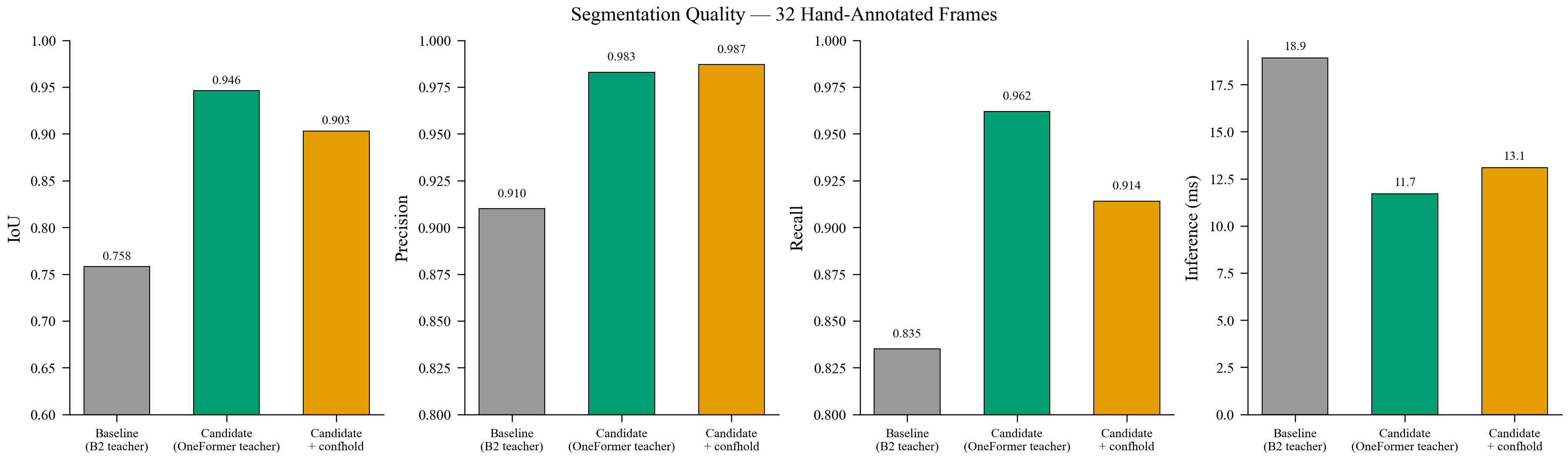}
  \caption{Segmentation quality comparison on 32 hand-annotated frames. The OneFormer-trained candidate improves all metrics while reducing inference time.}
  \label{fig:seg_improvement}
\end{figure}

Fig.~\ref{fig:seg_improvement} visualizes the per-metric improvement.
A full-video replay across six sequences (\numprint{22679} frames) confirms the improvement generalizes beyond the hand-annotated sample (Table~\ref{tab:fullvideo_replay}).

The candidate model reduces temporal instability by 77\% (1.46\%~$\to$~0.33\%) and increases template-path
availability by 5.6 percentage points.

\begin{table}[h]
\centering
\caption{Full-video replay metrics across six campus sequences (\numprint{22679} frames).}
\label{tab:fullvideo_replay}
\begin{tabular}{lccc}
\toprule
Metric & Baseline & Candidate & $\Delta$ \\
\midrule
Mean Seg IoU           & 0.909 & 0.925 & +0.016 \\
Unstable Rate [\%]     & 1.46  & 0.33  & $-$1.12\,pp \\
Has-Path Rate [\%]     & 100.0 & 100.0 & --- \\
Template Success [\%]  & 73.7  & 79.3  & +5.6\,pp \\
Fallback Rate [\%]     & 19.0  & 14.3  & $-$4.7\,pp \\
Mean Heading $\Delta$ [deg] & 0.209 & 0.201 & $-$0.008 \\
\bottomrule
\end{tabular}
\end{table}

\subsection{Planner Comparison Study}
\label{sec:planner_comparison}

\begin{figure}[!t]
\centering
\begin{subfigure}[b]{0.48\textwidth}
\includegraphics[width=\linewidth]{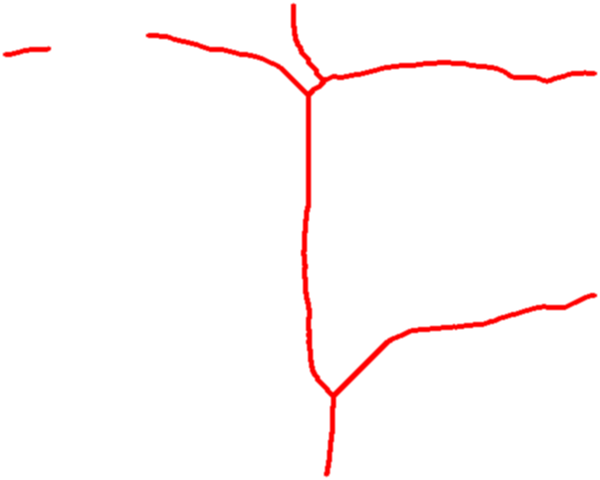}
\caption{Skeletonized BEV mask.}
\end{subfigure}\hfill
\begin{subfigure}[b]{0.48\textwidth}
\includegraphics[width=\linewidth]{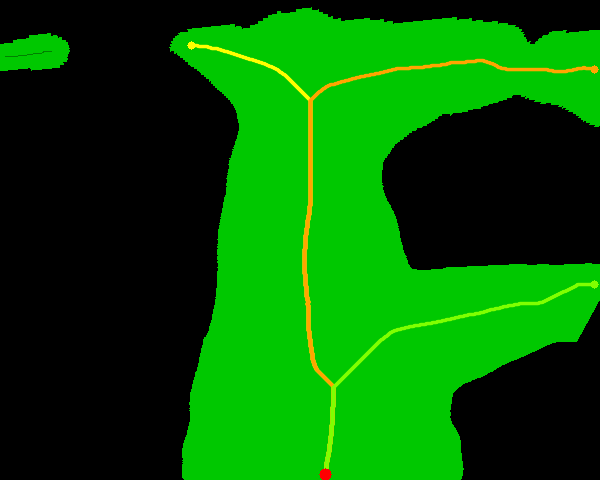}
\caption{Candidate paths on skeleton graph.}
\end{subfigure}

\medskip
\begin{subfigure}[b]{0.66\textwidth}
\includegraphics[width=\linewidth]{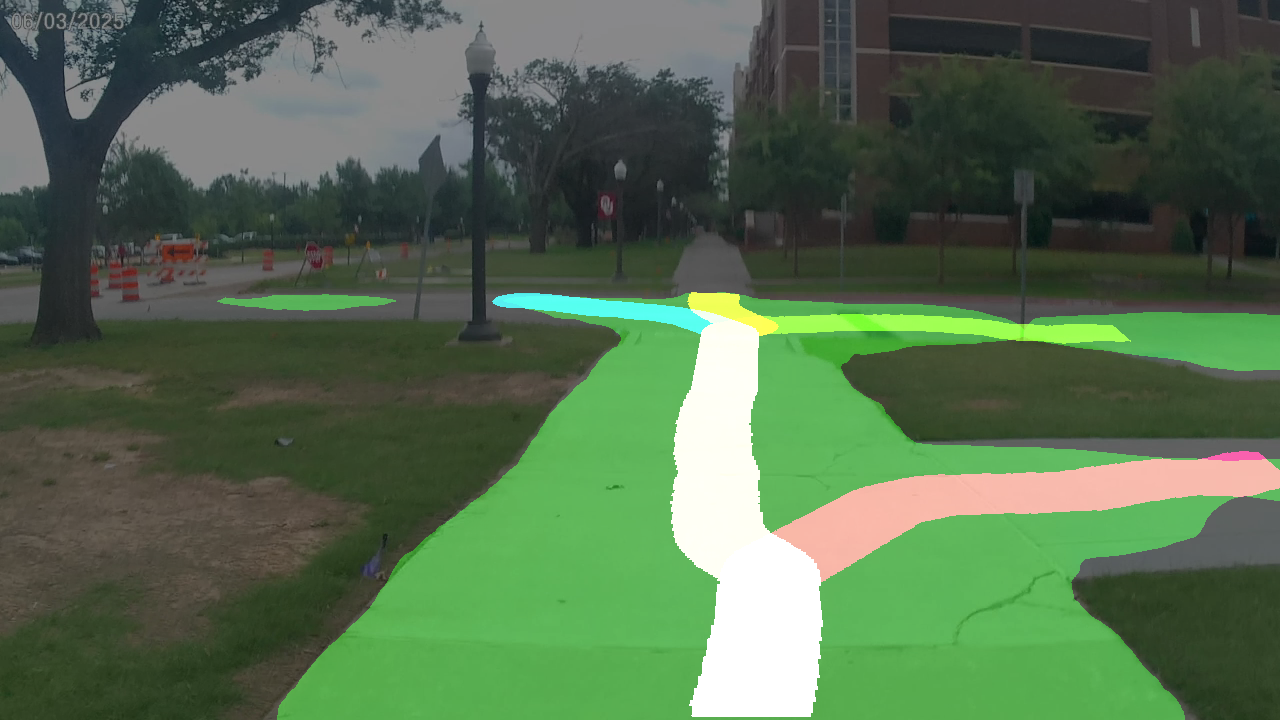}
\caption{Selected path reprojected to camera view.}
\end{subfigure}
\caption{BEV skeleton-graph pipeline (baseline method). (a)~Skeletonized BEV mask after Guo--Hall thinning, (b)~candidate paths from Dijkstra search, (c)~smoothed path reprojected to camera view. While topologically informative, this method is the slowest (\SI{380.3}{ms}) and least accurate of the five compared planners.}
\label{fig:skeleton_stage}
\end{figure}

We evaluate five planning methods on the 32 hand-annotated frames using the candidate segmentation model. Each planner receives the same cleaned binary mask; for BEV methods, the mask is first warped via homography (Fig.~\ref{fig:skeleton_stage} illustrates the BEV skeleton-graph pipeline as a representative example). Table~\ref{tab:planner_comparison} reports path availability, mask--path alignment (inside-GT ratio), lateral center error, and runtime.

\begin{table}[h]
\centering
\caption{Planner comparison on 32 hand-annotated frames (candidate mask). Runtime is mean per-frame wall time.}
\label{tab:planner_comparison}
\begin{tabular}{lp{0.7cm}ccc}
\toprule
Planner & Path [\%] & Inside-GT & Center Err. [px] & ms \\
\midrule
\multicolumn{5}{l}{\textit{BEV-domain methods}} \\
\quad Skeleton-Graph   & 100 & 0.971 & 76.6 & 380.3 \\
\quad DT Ridge (full)  & 100 & 0.986 & 65.0 & 926.8 \\
\quad DT Ridge (near)  & 100 & 0.986 & 79.0 & 1603.0 \\
\midrule
\multicolumn{5}{l}{\textit{Image-space methods}} \\
\quad Midpoint         & 100 & 0.985 & \textbf{14.3} & \textbf{2.2} \\
\quad DT Ridge         & 100 & \textbf{0.994} & 60.4 & 108.1 \\
\bottomrule
\end{tabular}
\end{table}

\begin{figure}[!t]
  \centering
  \includegraphics[width=\textwidth]{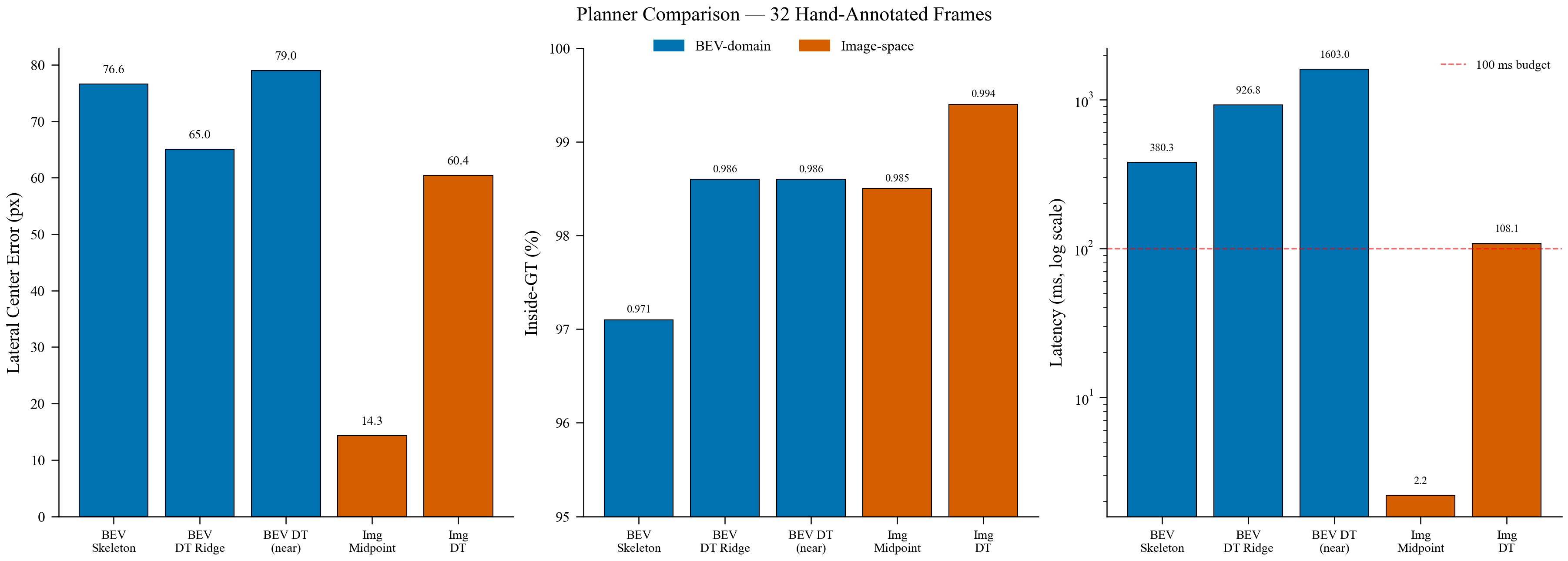}
  \caption{Planner comparison on 32 hand-annotated frames. Left: lateral center error (px) with runtime annotated. Right: inside-GT ratio. Image-space methods (green) dominate BEV methods (blue) in both accuracy and speed.}
\label{fig:planner_comparison}
\end{figure}

Fig.~\ref{fig:planner_comparison} visualizes the planner comparison results.

\paragraph{Key findings.}
\begin{enumerate}[nosep]
    \item \textbf{Image-space midpoint is the best primary planner.} It achieves the lowest lateral center error (14.3~px vs.\ 65.0~px for the best BEV method) at 421$\times$ lower latency (2.2~ms vs.\ 926.8~ms). Mask--path alignment is comparable (98.5\% vs.\ 98.6\%).
    \item \textbf{Image-space DT is the best fallback.} When midpoint extraction fails (discontinuous boundaries, irregular masks), image-space DT provides the highest inside-GT ratio (99.4\%) at \SI{108.1}{ms}---still 8.6$\times$ faster than BEV DT.
    \item \textbf{BEV methods are not competitive on runtime.} The skeleton-graph planner costs \SI{380.3}{ms} and produces the lowest alignment score. BEV DT costs \SI{926.8}{ms}. Near-field BEV DT (restricted to a 3\,m horizon) is actually \emph{slower} (\SI{1603}{ms}) due to denser graph construction in the near field.
    \item \textbf{BEV does not improve path quality.} Even with oracle (ground-truth) masks, BEV DT achieves only 98.0\% inside-GT ratio and 69.2~px center error, compared to 98.9\% and 62.5~px for image-space DT and 98.4\% and 15.2~px for image-space midpoint. The planning \emph{domain} is part of the problem, not just the segmentation input.
\end{enumerate}

\subsection{BEV Fragility Analysis}
\label{sec:bev_fragility}

Beyond the latency disadvantage, BEV-only planning exhibits a critical reliability failure in monocular settings. In a profiled run of 4,407 frames with the baseline segmentation model and full BEV pipeline, Table~\ref{tab:bev_fragility} shows that 99.3\% of frames produced no valid BEV path.

\begin{table}[h]
\centering
\caption{BEV path extraction reliability over 4,407 profiled frames.}
\label{tab:bev_fragility}
\begin{tabular}{lr}
\toprule
Metric & Value \\
\midrule
Frames with valid path (\texttt{dt\_ridge}) & 10 (0.2\%) \\
Frames with held path (\texttt{dt\_ridge\_hold}) & 20 (0.5\%) \\
Frames with no path (\texttt{none}) & 4,377 (99.3\%) \\
Mean BEV mask occupancy ratio & 0.0002 \\
\bottomrule
\end{tabular}
\end{table}

\begin{figure}[!t]
  \centering
  \includegraphics[width=0.72\textwidth]{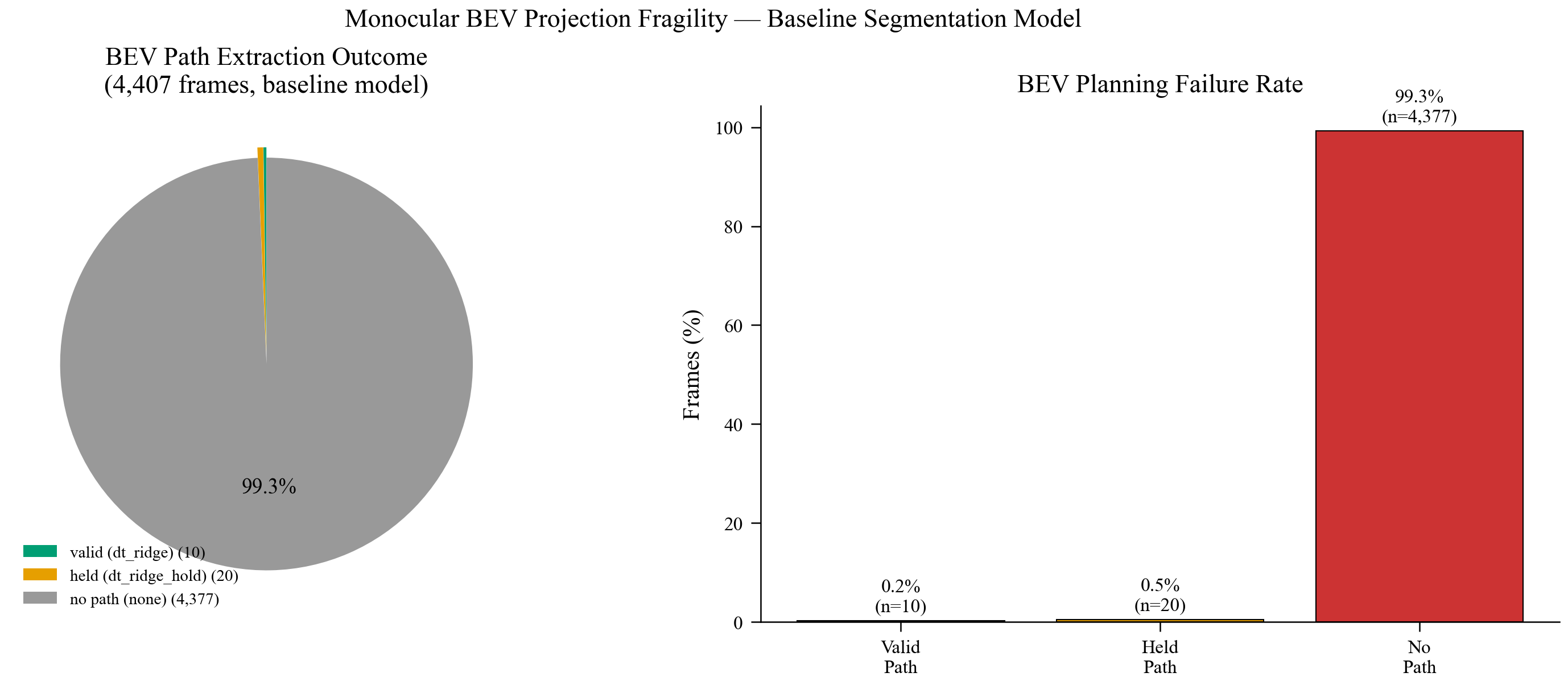}
  \caption{BEV path extraction outcome over 4,407 frames. 99.3\% of frames produced no valid BEV path, demonstrating the fragility of monocular BEV planning.}
\label{fig:bev_fragility}
\end{figure}

Fig.~\ref{fig:bev_fragility} visualizes this distribution. The root cause is that monocular perspective projection maps a narrow forward-view strip into the BEV grid, leaving most of the BEV frame empty. The distance transform and skeleton algorithms then operate on a tiny sliver of occupied pixels, frequently failing to produce a connected path of sufficient length. This fragility is fundamental to the monocular BEV geometry---it cannot be resolved by better segmentation alone, as the oracle-mask experiments confirm. This finding motivates our recommendation to use image-space planning as the primary domain.

\subsection{System Runtime}
\label{sec:res_runtime}

\begin{figure}[!t]
  \centering
  \includegraphics[width=0.88\textwidth]{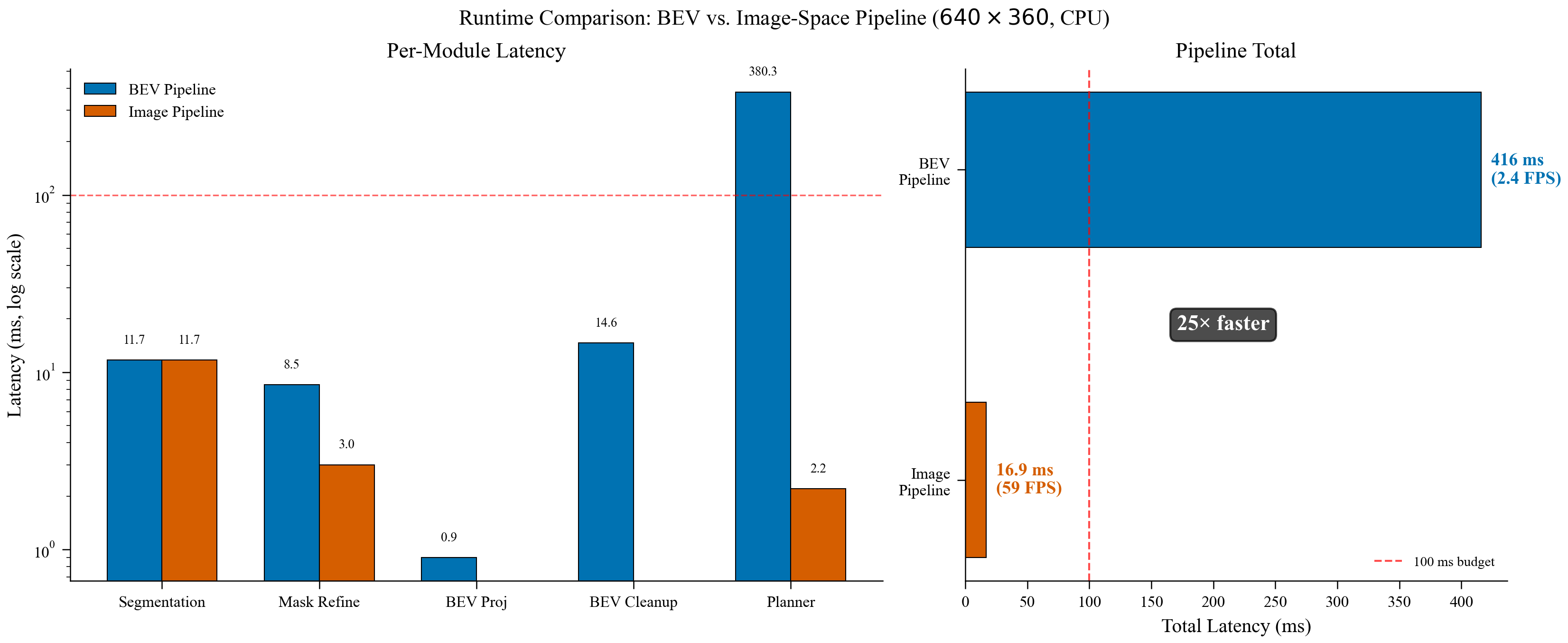}
  \caption{Per-module runtime comparison. The image-space architecture eliminates BEV overhead and reduces planner cost by 173$\times$, achieving 59~FPS vs.\ 2.4~FPS.}
\label{fig:runtime_breakdown}
\end{figure}

Table~\ref{tab:runtime_comparison} and Fig.~\ref{fig:runtime_breakdown} compare per-module runtimes for the BEV pipeline (skeleton planner) and the recommended image-space pipeline (midpoint primary, DT fallback). The image-space architecture eliminates BEV warp and runs the planner in \SI{2.2}{ms} instead of \SI{380.3}{ms}, achieving over 59~FPS on CPU---more than sufficient for a pedestrian-speed platform.

\begin{table}[h]
\centering
\caption{Per-module runtime comparison at $640{\times}360$ (CPU-only).}
\label{tab:runtime_comparison}
\begin{threeparttable}
\begin{tabular}{lcc}
\toprule
Module & BEV Skeleton & Image-Space \\
\midrule
SegFormer Inference   & 11.7\,ms & 11.7\,ms \\
Mask Refinement       & 8.5\,ms & 3.0\,ms\tnote{$\dagger$} \\
BEV Projection        & 0.9\,ms & ---      \\
BEV Cleanup           & 14.6\,ms & ---     \\
Planner               & 380.3\,ms & 2.2\,ms \\
\midrule
Total                 & 416.0\,ms & 16.9\,ms \\
FPS                   & 2.4       & \textbf{59.2}  \\
\bottomrule
\end{tabular}
\begin{tablenotes}
\small
\item[$\dagger$] Image-space mode skips BEV-specific cleanup stages.
\end{tablenotes}
\end{threeparttable}
\end{table}

Table~\ref{tab:runtime_offenders} ranks the major runtime offenders, guiding deployment optimization. The dominant cost in the BEV pipeline is the planner itself, not display or I/O: a flag sweep over GUI/headless and save/no-save modes showed less than \SI{5}{ms} variation, while BEV DT planning alone consumed \SI{926.8}{ms} per frame.

\begin{table}[h]
\centering
\caption{Runtime offenders ranked by cost. The image-space architecture eliminates the top three.}
\label{tab:runtime_offenders}
\begin{tabular}{lrl}
\toprule
Component & Cost (ms) & Recommendation \\
\midrule
BEV DT Planner        & 926.8 & Replace \\
BEV Graph Planner      & 380.3 & Replace \\
YOLO Detection (CPU)   & 39.0  & Gate or GPU \\
BEV Warp + Cleanup     & 15.5  & Skip in img mode \\
Seg ($512{\times}288$) & 75.3  & Use $640{\times}360$ \\
Predictor disabled     & +73.0 & Always enable \\
\bottomrule
\end{tabular}
\end{table}

Table~\ref{tab:runtime_configs} shows profiled runtimes under different system configurations, demonstrating the impact of detection mode and the BEV predictor (skip-frame reuse) on overall throughput.

\begin{table}[h]
\centering
\caption{System-level runtime under different configurations (4,407-frame profiled run, $512{\times}288$ input).}
\label{tab:runtime_configs}
\begin{tabular}{lcccc}
\toprule
Configuration & FPS & Seg (ms) & Det (ms) & BEV (ms) \\
\midrule
No detection, predictor on    & 25.3 & 20.3 & --- & 11.0 \\
GPU detection, predictor on   & 20.1 & 17.0 & 12.2 & 13.4 \\
CPU detection, predictor on   & 12.2 & 18.8 & 39.0 & 14.6 \\
No detection, predictor off   & 8.9  & 75.3 & --- & 29.5 \\
\bottomrule
\end{tabular}
\end{table}

\subsection{Failure Modes and Limitations}
\label{sec:failure_modes}

While the system performs robustly on most campus paths, we observed failures in specific scenarios:

\begin{itemize}[noitemsep,topsep=2pt]
    \item \textbf{Wide intersections or plazas:} At large open junctions with weak edge cues, the segmentation may fail to resolve sidewalk continuation, causing incorrect path selection in all planners.
    \item \textbf{Partial occlusions:} Pedestrians or parked objects create temporary gaps in the mask. Image-space planners handle this more gracefully than BEV planners because the gap remains geometrically coherent in image space but can produce disconnected fragments in BEV.
    \item \textbf{Junction ambiguity:} In a few scenes, the ``natural'' forward direction is difficult to define; human labelers disagreed.
    \item \textbf{Edge bleeding:} The student model occasionally overshoots sidewalk boundaries in bright or low-contrast regions, though the improved model reduces this substantially (precision 0.910~$\to$~0.983).
    \item \textbf{BEV coverage:} As documented in Section~\ref{sec:bev_fragility}, the monocular viewpoint provides insufficient mask coverage for reliable BEV-domain planning.
\end{itemize}

Dynamic obstacle handling is limited to implicit mask avoidance: moving objects appear as non-traversable regions, and the path routes around them when possible. Explicit obstacle detection and prediction are deferred to future work.

\section{Conclusion}
\label{sec:conclusion}

We presented a modular vision-based sidewalk navigation pipeline that evolves through three design iterations, from a skeleton-graph baseline through distance-transform corridor planning to a lightweight image-space architecture. Through systematic comparison of five planning methods, we arrive at a result that is both an engineering improvement and a stronger scientific contribution: \emph{for monocular sidewalk navigation on embedded platforms, simple image-space geometry consistently outperforms the more complex BEV pipeline in accuracy, reliability, and latency.}

The key quantitative findings are:
\begin{itemize}[nosep]
    \item The OneFormer-trained SegFormer-B0 student achieves hand-annotated IoU of 0.946 at \SI{11.7}{ms}, improving over the baseline (0.758, \SI{18.9}{ms}).
    \item Image-space midpoint planning achieves 14.3~px lateral center error at \SI{2.2}{ms}---421$\times$ faster than BEV DT planning (65.0~px, \SI{926.8}{ms}).
    \item BEV-only planning fails on 99.3\% of frames in one profiled sequence due to insufficient monocular mask coverage.
    \item The recommended image-space architecture runs the full perception-to-path stack at over 59~FPS on CPU.
\end{itemize}

The final recommended architecture---image-space midpoint primary, image-space DT fallback, BEV reserved for optional visualization and obstacle projection---is transparent, lightweight, and deployable on single-board computers. Each stage produces an interpretable intermediate representation that field teams can inspect and tune without retraining models.

Several limitations remain. All evaluations are conducted offline without closing the loop on the physical scooter. Dynamic obstacles are not modeled explicitly. The homography is fixed for a single camera pose. The planner comparison was conducted on 32 hand-annotated frames, and while full-video replay confirms generalization, a larger annotated benchmark would strengthen the findings.

Future work will extend the system in three directions. First, closed-loop deployment on the scooter platform with real-time obstacle detection and avoidance. Second, temporal video-based segmentation to further improve mask stability. Third, coupling the local image-space planner with global route planning and mapping for longer autonomous traversals.

\section*{Funding}
This research did not receive any specific grant from funding agencies in the
public, commercial, or not-for-profit sectors.

\section*{Declaration of generative AI and AI-assisted technologies in the manuscript preparation process}
During the preparation of this work, the authors used Claude (Anthropic) to assist
with code development, data analysis, and language editing. After using these
tools, the authors reviewed and edited the content as needed and take full
responsibility for the content of this paper.

\section*{Data availability}
Original code and configuration files for the navigation pipeline are available in
the ScooterProject repository at
\url{https://github.com/Lkhanaajav/live_test_scooter_project}. The raw campus video
recordings contain identifiable pedestrians and cannot be shared publicly due to
privacy considerations. Aggregated evaluation logs and example anonymized
frames are available from the corresponding author on reasonable request.

\section*{CRediT authorship contribution statement}
L. Mijiddorj: Conceptualization, Methodology, Software, Validation,
Formal analysis, Investigation, Data curation, Visualization,
Writing -- original draft, Writing -- review \& editing.
B. Mijiddorj: Conceptualization, Investigation, Writing -- review \& editing.
Y. Yan: Conceptualization, Data curation, Writing -- review \& editing.
A. Ho: Conceptualization, Writing -- review \& editing.
T. Beringer: Conceptualization, Investigation, Writing -- review \& editing.
B. Xu: Conceptualization, Supervision, Writing -- review \& editing.
B. Weng: Conceptualization, Supervision, Project administration,
Funding acquisition, Resources, Writing -- review \& editing.

\bibliographystyle{IEEEtran}
\bibliography{cas-refs}

\end{document}